\pdftrailerid{}
\documentclass[letterpaper]{article}
\usepackage[preprint]{aaai2027}
\usepackage[hyphens]{url}
\usepackage{graphicx}
\usepackage{natbib}
\usepackage{caption}
\usepackage{booktabs}
\usepackage{amsmath}
\usepackage{amsfonts}
\usepackage{nicefrac}
\usepackage{multirow}
\usepackage{makecell}
\usepackage{array}
\usepackage{algorithm}
\usepackage{algpseudocode}
\title{FadeMem: Distance-Aware Memory Consolidation\\ for Autoregressive Video Diffusion}
\author{
Yu Lu\textsuperscript{1,*},\quad
Junjie Yang\textsuperscript{1,*},\quad
Piotr Koniusz\textsuperscript{2,3},\quad
YuXin Song\textsuperscript{4},\quad
Yi Yang\textsuperscript{1}
}
\affiliations{
\textsuperscript{1}Zhejiang University \quad
\textsuperscript{2}University of New South Wales (UNSW) \quad
\textsuperscript{3}Data61/CSIRO \quad
\textsuperscript{4}Baidu Inc.\\
\textbf{Project page:} \url{https://fademem.github.io/FadeMem_Webpage/}
}
\begin{document}
\maketitle
\begingroup
\renewcommand{\thefootnote}{*}
\footnotetext{Equal contribution.}
\endgroup
\begin{abstract}
Autoregressive video generators synthesize long videos by generating successive temporal segments, but their historical KV cache grows with video length.
Existing bounded-cache methods reduce this cost with local windows, sink tokens, or compressed memory states, yet they usually assign fixed roles to different parts of the history.
We propose \textbf{FadeMem}, a distance-aware KV memory consolidation mechanism that organizes historical KV blocks into a temporal hierarchy under a fixed cache budget.
This design is motivated by frequency-dependent temporal decay: fine details decorrelate quickly, while coarse scene structure and identity remain useful over longer horizons.
During generation, new history is inserted as fine-grained entries, while older adjacent entries are progressively merged under a power-law temporal allocation schedule, yielding a dense-near, sparse-far memory within one cache.
Without architectural changes, FadeMem improves long-range consistency while largely preserving visual quality, and lightweight adaptation further enhances motion dynamics and visual fidelity. Using the same unified schedule under a fixed cache budget, FadeMem also remains effective over multi-minute and hour-long video generation and reduces peak memory under a matched KV budget.

\end{abstract}

\section{Introduction}
Recent advances in diffusion and transformer-based video generation have substantially improved the visual quality, controllability, and semantic fidelity of text-to-video synthesis and editing~\cite{ho2020ddpm,song2020ddim,song2021score,rombach2022ldm,peebles2023dit,ho2022videodiffusion,ho2022imagenvideo,singer2022makeavideo,blattmann2023videoldm,blattmann2023stablevideodiffusion,kondratyuk2023videopoet,yang2024cogvideox,kong2024hunyuanvideo,polyak2024moviegen,wan2025wan,zhang2026sama}.
More recently, autoregressive video generators have extended this progress to long-horizon video synthesis by generating videos sequentially over time~\cite{causvid,selfforcing,longlive}.
In this sequential setting, previously generated frames provide the temporal context needed to maintain subject appearance, scene layout, and motion continuity over long horizons.
The key challenge is therefore how to retain and use this continuously growing history as generation proceeds.
A naive method is to cache all historical context as KV blocks.
However, its storage overhead and attention computation grow with video length, making it impractical for long-video generation.

\begin{figure*}[t]
    \centering
    \includegraphics[width=0.98\textwidth]{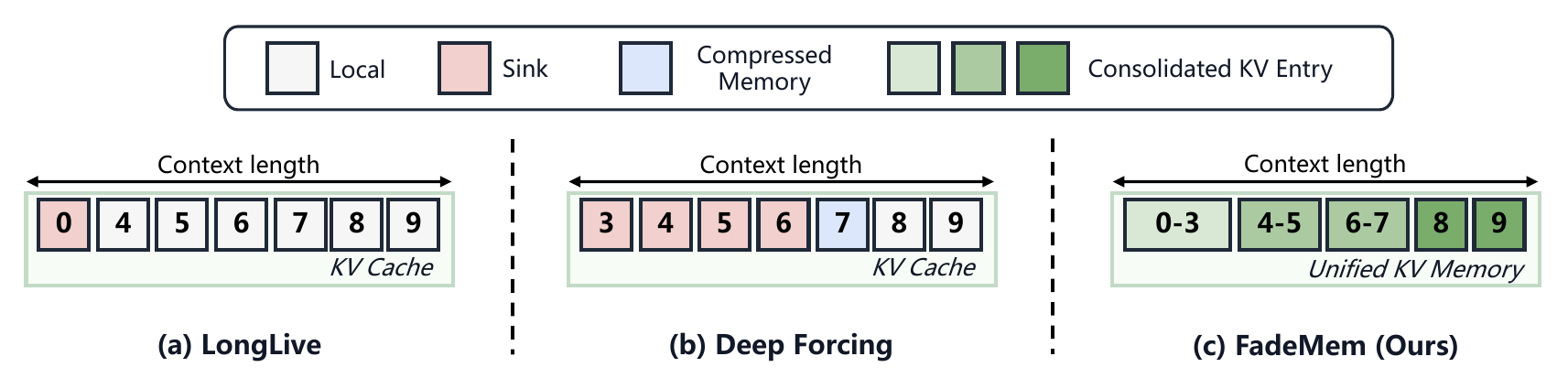}
    \caption{
    Bounded KV cache structures for long-horizon video generation.
    LongLive keeps sink tokens and a local window, Deep Forcing adds compressed memory states, and FadeMem organizes a single bounded cache into a temporal hierarchy that preserves recent history at fine temporal resolution while consolidating distant history into entries that summarize longer temporal spans.
    }
    \label{fig:kv_cache_comparison}
\end{figure*}

Existing methods reduce this cost by retaining only selected portions of the history, such as a local window of recent frames, persistent sink tokens from the beginning of the video, or a small number of compressed memory states~\cite{streamingllm,longlive,deepforcing,memrope}~(as shown in Figure~\ref{fig:kv_cache_comparison}).
Although effective, these designs typically rely on manually specified cache partitions with fixed temporal roles, rather than allocating memory adaptively according to temporal distance.

In this work, we revisit bounded KV cache management from the perspective of temporal distance.
To motivate this view, we analyze how frame-to-frame correlations evolve across different temporal lags in long videos.
As shown in Figure~\ref{fig:observation}, correlations generally decrease as the temporal distance between frames increases.
Importantly, this decay is frequency-dependent: high-frequency details, such as fine textures, local motion, and small appearance variations, decorrelate rapidly, whereas lower-frequency structures, such as scene layout, object identity, and global appearance, remain correlated over longer horizons.
This suggests a distance-dependent role of historical context: nearby history is crucial for local motion continuity and short-term visual consistency, while distant history primarily provides long-range structural anchors.
Together, these observations imply that historical context should be represented at a temporal resolution that decreases with distance: nearby states should remain fine-grained to preserve local dynamics, whereas distant history can be summarized over longer temporal spans.

Based on this principle, we propose \textbf{FadeMem}, a distance-aware memory consolidation mechanism for autoregressive long-video generation.
FadeMem maintains a single bounded memory composed of entries, each storing a representative historical KV block together with the temporal span it summarizes.
A power-law temporal allocation schedule controls how the memory resolution changes with temporal distance.
Each new KV block is initially stored as an individual entry; when the fixed entry budget is exceeded, the schedule determines which temporally adjacent entries to consolidate into a single entry representing their combined temporal span.
Repeating this online update keeps recent history at fine temporal resolution while progressively summarizing distant history over longer spans.
FadeMem thereby covers an increasingly long history with a fixed number of entries, without separate local, sink, or auxiliary memory modules.
The resulting memory hierarchy preserves recent context for short-term dynamics while retaining compact long-range anchors for scene layout, appearance, and identity.
FadeMem requires no architectural modification and supports both inference-time use and light fine-tuning.

Without retraining, FadeMem improves long-range subject, background, and temporal consistency while largely preserving the visual quality of the base generator. With lightweight adaptation, the model further learns to exploit consolidated historical states, leading to marked improvements in motion dynamics and visual quality. FadeMem also remains effective over four-minute and hour-long video generation and reduces peak memory under a matched KV budget.

Our contributions are summarized as follows:
\begin{itemize}
    \item We provide an empirical analysis of distance-dependent spectral decay in long videos, showing that fine-grained details decorrelate faster than coarse scene-level structure over long temporal distances.
    \item We propose FadeMem, a unified memory mechanism for cached KV blocks that forms a temporal hierarchy of memory spans under a fixed cache budget.
    \item We demonstrate that FadeMem improves long-horizon video coherence in fixed-budget autoregressive generation, effectively reducing identity drift and scene degradation under bounded memory.
\end{itemize}

\section{Related Work}
\subsection{Autoregressive Video Generation}
Autoregressive video generation has emerged as an effective paradigm for long-horizon and streaming video synthesis.
Recent methods extend diffusion generation through long-context modeling, streaming generation, or next-token-style prediction~\cite{diffusionforcing,streamingt2v,streamdit,ltxvideo,inflvg,skyreelsv2,magi1,framepack}.
Training-free spectral attention methods have also explored long video generation by manipulating temporal attention in the frequency domain~\cite{lufreelong,lu2025freelongpp}.
CausVid~\cite{causvid} distills bidirectional video diffusion into a causal generator with KV cache reuse, while Self Forcing~\cite{selfforcing}, Self-Forcing++~\cite{selfforcingpp}, Rolling Forcing~\cite{rollingforcing}, and LongLive~\cite{longlive} improve long rollouts through self-generated histories, error correction, rolling denoising windows, or streaming-oriented tuning.
FAR~\cite{far} further studies longer-context autoregressive video modeling with distant-frame compression and positional extrapolation.
These methods advance long-form generation, but their temporal consistency still depends on how historical information is retained under a bounded cache budget.
FadeMem focuses on this cache organization problem.
\subsection{Memory in Long Video Generation}
A common way to control memory growth is to keep a sliding local window and optionally preserve early frames or sink tokens as persistent anchors~\cite{longlive,streamingllm}.
Infinity-RoPE~\cite{infinityrope} addresses positional issues when cached context is reused over long rollouts.
Other methods introduce more active memory mechanisms, including attention-based token compression~\cite{deepforcing}, long- and short-term memory streams~\cite{memrope}, adaptive memory retrieval~\cite{memflow}, and long-context organization for extended generation~\cite{contextasmemory,mixturecontexts}.
Related LLM cache-compression methods also study token retention under fixed KV budgets~\cite{li2024snapkv,cai2024pyramidkv}.
In contrast, FadeMem treats bounded-cache management as temporal-resolution allocation.
Rather than assigning fixed roles to cache slots or selecting individual tokens, it progressively consolidates older KV blocks into span-level entries according to temporal distance, preserving dense recent context while maintaining compact distant coverage.

\section{Method}
\label{sec:method}
\begin{figure*}[!t]
    \centering
    \includegraphics[width=0.98\textwidth]{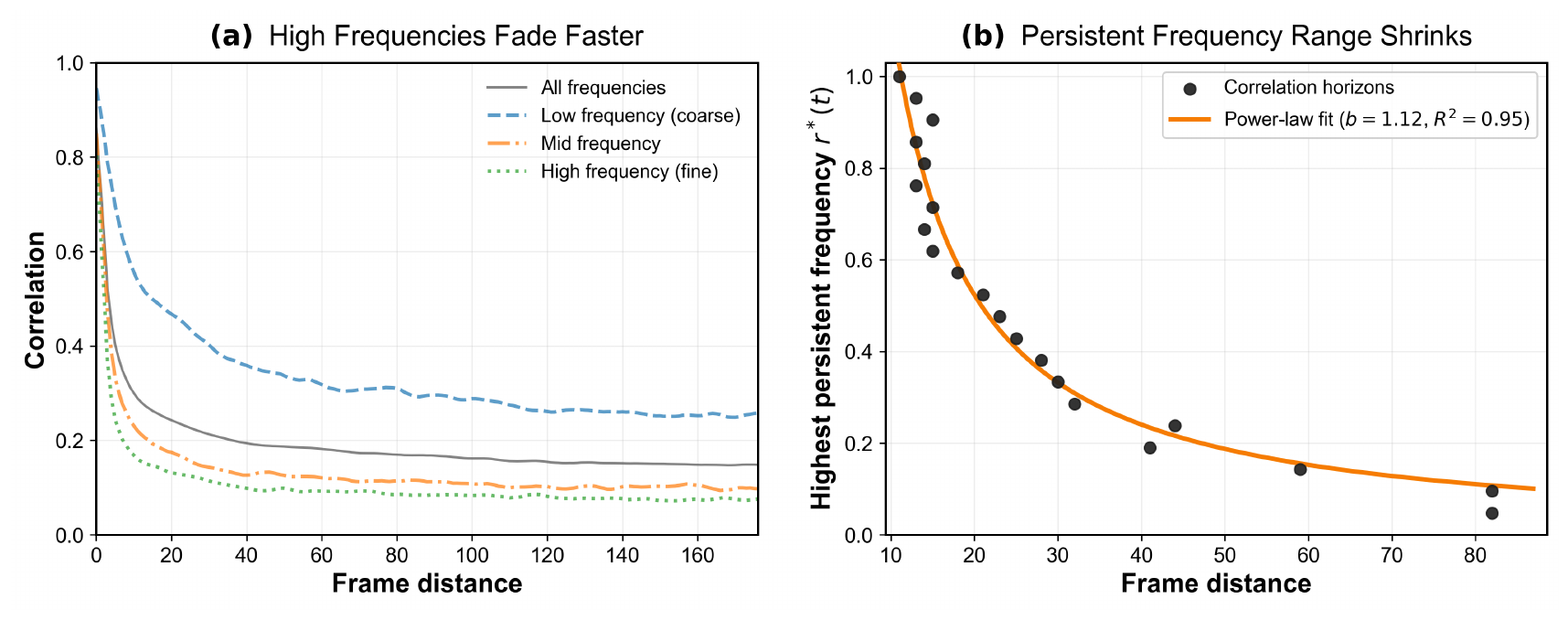}
    \caption{Temporal information becomes progressively low-pass as history recedes.
(a) Correlations between latent representations decrease with frame distance, with high-frequency components corresponding to fine appearance and local motion decaying substantially faster than low-frequency scene and identity structure.
(b) Using the decorrelation horizon of each frequency band, we estimate $r^{*}(t)$, the highest frequency that remains temporally persistent at distance t. Its approximate power-law decay indicates that recent history should be retained at fine granularity, whereas distant history can be consolidated into coarser long-range anchors.}
    \label{fig:observation}
\end{figure*}
\begin{figure*}[t]
    \centering
    \includegraphics[width=0.98\textwidth]{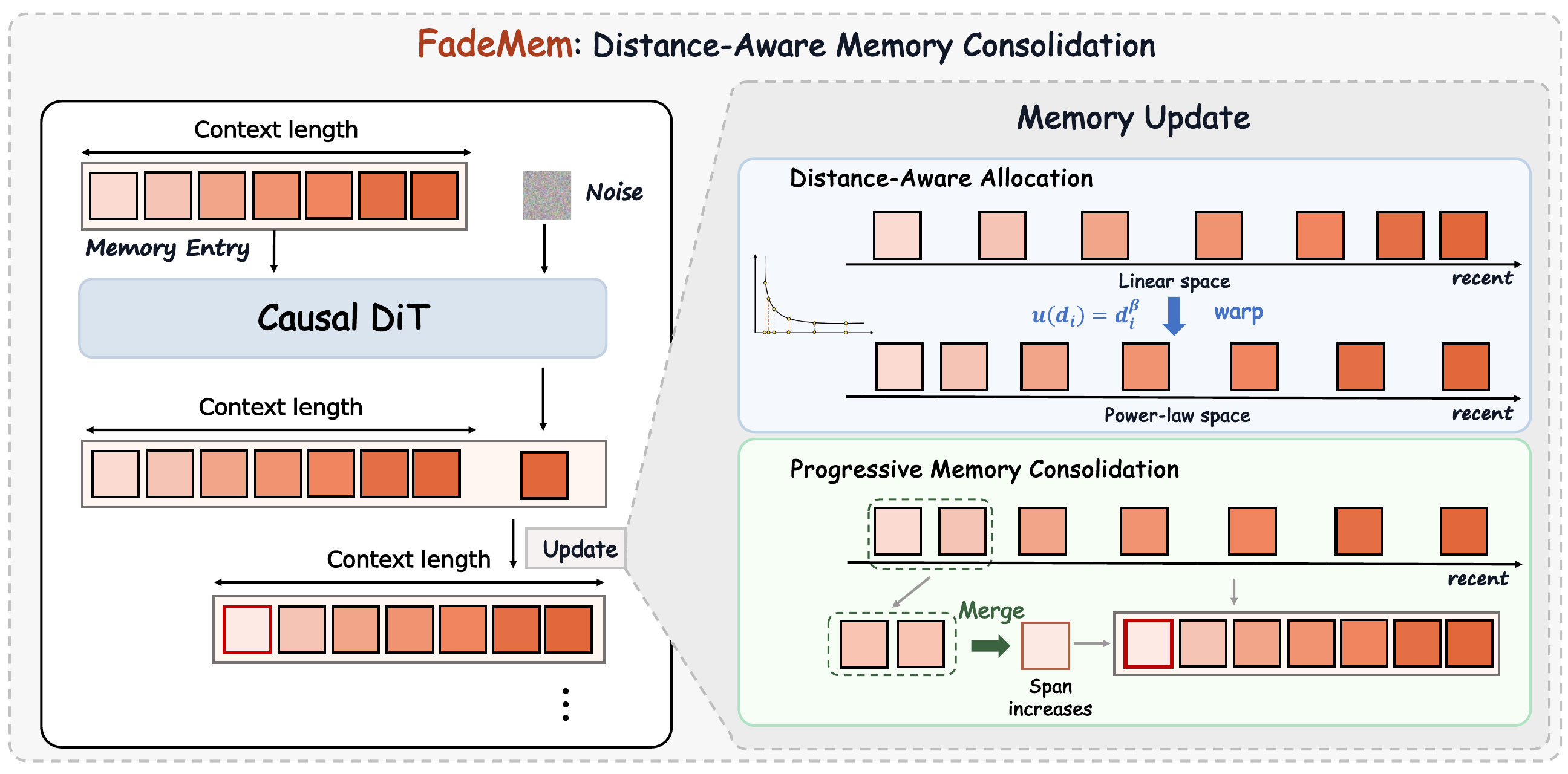}
    \caption{Overview of FadeMem.
    FadeMem organizes KV entries as a temporal hierarchy, keeping recent entries fine-grained while progressively merging older adjacent entries into coarser summaries under a fixed cache budget.}
    \label{fig:method_overview}
\end{figure*}
FadeMem formulates bounded KV cache management as distance-aware consolidation over generated history.
As temporal distance increases, frame-level correlations decay and the remaining reliable information becomes increasingly structural.
FadeMem captures this pattern with a temporal allocation schedule that keeps nearby KV blocks fine-grained and progressively consolidates distant blocks within a bounded memory, as shown in Figure~\ref{fig:method_overview}.
\subsection{Problem Formulation}
\label{sec:formulation}
Consider an autoregressive video generator that produces a sequence of generation units, such as frames, latent frames, short chunks, or token blocks depending on the backbone.
Let $x_{1:t}$ denote the generated history before predicting $x_{t+1}$.
During inference, each transformer layer stores the key-value states of previous units in a KV cache.
We call the keys and values produced by one unit a KV block and denote the block at step $\tau$ as
\[
\mathbf{B}_{\tau} = (\mathbf{K}_{\tau}, \mathbf{V}_{\tau}).
\]
A full-history cache stores all previous blocks $\{\mathbf{B}_{\tau}\}_{\tau=1}^{t}$, providing complete context at a memory cost that grows linearly with video length.
FadeMem replaces the full-history cache with a bounded memory $\mathcal{M}_t$ with a fixed entry budget $M$:
\[
|\mathcal{M}_t| \leq M.
\]
The memory is an ordered set of entries.
Each entry stores one representative KV block and temporal metadata, such as its summarized span.
If each generation unit contains $N$ visual tokens, each memory entry stores $N$ token-level key/value vectors per layer.
With at most $M$ entries, FadeMem stores at most $MN$ token-level key/value vectors per layer while representing an increasingly long history.
\subsection{Distance-Dependent Spectral Decay}
\label{sec:spectral_decay}
We examine how temporal correlation changes with distance across frequency bands.
Given a video, we encode each frame into a latent representation, decompose each latent frame into radial frequency bands, and compute the correlation $C_r(\Delta)$ between latent coefficients in band $r$ separated by temporal lag $\Delta$.
As shown in Figure~\ref{fig:observation}(a), correlations decrease with lag, and the decay is frequency-dependent.
High-frequency components, such as fine texture, local motion, and small appearance variations, decorrelate rapidly.
Low-frequency components, such as scene layout, object identity, and global appearance, remain correlated longer.
We summarize this trend with a decorrelation horizon $t^*(r)$ for each band, defined as the temporal lag where the correlation curve enters a slowly varying, near-flat regime.
The boundary points $(t^*(r), r)$ indicate which frequencies remain stable up to each distance.
Equivalently, for temporal distance $t$, we define the stable frequency radius $r^*(t)$ as the largest radial frequency band whose decorrelation horizon is at least $t$.
As shown in Figure~\ref{fig:observation}(b), $r^*(t)$ decreases with temporal distance, indicating that distant history mainly preserves low-frequency structure.
We observe that the boundary follows an approximate power-law trend:
\begin{equation}
r^*(t) \propto t^{-b},
\end{equation}
where $b$ controls how quickly the stable frequency bandwidth decays over time.
This pattern suggests that memory density should decrease with temporal distance: recent states need fine temporal granularity, while distant states can be represented more coarsely.
\subsection{Distance-Aware Memory Consolidation}
\label{sec:temporal_allocation}
FadeMem implements this dense-near, sparse-far principle with a single ordered memory, rather than separate local, sink, or auxiliary cache components.
\paragraph{Memory entries and insertion.}
At generation step $t$, the memory is an ordered sequence of entries:
\[
\mathcal{M}_t
=
\{m_1^t, m_2^t, \ldots, m_{M_t}^t\},
\quad
M_t \leq M.
\]
Each entry stores a KV block together with lightweight temporal metadata:
\[
m_i^t
=
\left(
\bar{\mathbf{K}}_i^t,
\bar{\mathbf{V}}_i^t,
\mu_i^t,
s_i^t
\right),
\]
where $\bar{\mathbf{K}}_i^t$ and $\bar{\mathbf{V}}_i^t$ are the stored key and value states, $\mu_i^t$ is the temporal position represented by the entry, and $s_i^t$ is the number of generation units summarized by the entry.
A new entry has $s_i^t=1$; a consolidated entry may represent a longer segment.
After generating a new unit $x_t$, FadeMem inserts its KV block as a new entry:
\[
m_{\mathrm{new}}^t
=
\left(
\mathbf{K}_{t},
\mathbf{V}_{t},
t,
1
\right).
\]
If the number of entries remains within the budget $M$, the new entry is simply appended.
Otherwise, FadeMem consolidates one adjacent pair.
\paragraph{Distance-aware scheduling.}
For each eligible entry, we measure its temporal distance from the current generation step,
\[
d_i = t - \mu_i,
\]
and map this distance into a warped temporal space:
\[
u(d_i) = d_i^{\beta},
\quad 0 < \beta \leq 1,
\]
where $\beta$ controls the strength of temporal compression.
This power-law mapping keeps nearby entries relatively separated while bringing distant entries closer together.
FadeMem selects the adjacent pair with the smallest gap in warped space:
\[
j
=
\arg\min_i
\left|
u(d_{i+1}) - u(d_i)
\right|.
\]
Because only adjacent entries are merged, the memory remains ordered.
Over time, the ordered memory becomes a temporal hierarchy of variable-span entries: recent entries keep fine temporal granularity, while entries farther in the past cover progressively longer intervals.


\paragraph{Consolidation operator.}
For the selected entries $m_i$ and $m_{i+1}$, FadeMem first merges their temporal metadata:
\begin{align*}
s_{\mathrm{new}} &= s_i + s_{i+1}, \\
\mu_{\mathrm{new}} &=
\frac{s_i \mu_i + s_{i+1} \mu_{i+1}}
{s_i + s_{i+1}}.
\end{align*}
The new span records the total covered length, and the new center gives the span-weighted temporal position.
The stored keys and values are then merged into a representative KV block:
\[
(\bar{\mathbf{K}}_{\mathrm{new}}, \bar{\mathbf{V}}_{\mathrm{new}})
=
\operatorname{Merge}(m_i,m_{i+1}).
\]
The resulting entry replaces the selected pair, keeping the memory size fixed.
By default, we use fixed square-root span weighting to prevent increasingly long-range summaries from dominating later consolidations:
\begin{align*}
\bar{\mathbf{K}}_{\mathrm{new}}
&=
\frac{
\sqrt{s_i}\,\bar{\mathbf{K}}_i
+
\sqrt{s_{i+1}}\,\bar{\mathbf{K}}_{i+1}
}{
\sqrt{s_i}+\sqrt{s_{i+1}}
}, \\
\bar{\mathbf{V}}_{\mathrm{new}}
&=
\frac{
\sqrt{s_i}\,\bar{\mathbf{V}}_i
+
\sqrt{s_{i+1}}\,\bar{\mathbf{V}}_{i+1}
}{
\sqrt{s_i}+\sqrt{s_{i+1}}
}.
\end{align*}
Other local operators, such as unweighted averaging or representative selection, are compared in Table~\ref{tab:ablation_consolidation}.
The update is online and local: each generation step adds one entry and, when necessary, performs one consolidation.


\paragraph{Boundary and positional handling.}
FadeMem uses simple boundary handling for important temporal references.
The newest entry is protected from immediate consolidation, so the latest KV block remains available for at least one update step.
By default, FadeMem preserves the first-frame KV block as a global anchor within the same ordered consolidation schedule.
The complete memory-update pseudocode is provided in the supplementary material.
Because RoPE~\cite{su2021roformer} encodes temporal position into key states, directly merging RoPE-encoded keys may mix incompatible positional phases.
FadeMem therefore stores memory keys after removing their original temporal RoPE phase.
Consistent with prior inference-time memory methods that adjust temporal RoPE indices~\cite{deepforcing,memrope}, we re-index retained memory keys within a bounded temporal range at attention time.
Thus, FadeMem remains compatible with RoPE-based attention while keeping at most $M$ entries per layer and changing only the cached context exposed to the backbone.

\section{Experiments}
\label{sec:experiments}
\begin{table*}[t!]
\centering
\small
\setlength{\tabcolsep}{1.5pt}
\renewcommand{\arraystretch}{1.10}
\begin{tabular}{lccccccc}
\toprule
\multirow{2}{*}{\textbf{Method}} &
\multicolumn{7}{c}{\textbf{Evaluation Scores} $\uparrow$} \\
\cmidrule(lr){2-8}
&
\makecell[c]{Subject\\Consistency} &
\makecell[c]{Background\\Consistency} &
\makecell[c]{Motion\\Smoothness} &
\makecell[c]{Dynamic\\Degree} &
\makecell[c]{Aesthetic\\Quality} &
\makecell[c]{Imaging\\Quality} &
\makecell[c]{Avg.} \\
\midrule
\multicolumn{8}{l}{\textit{Baselines and inference-time memory variants}} \\
Self Forcing~\cite{selfforcing}
& 97.08 & 96.32 & 98.37 & 33.88 & 56.87 & 66.92 & 78.64 \\
MemFlow~\cite{memflow}
& 97.44 & 96.27 & 98.72 & 40.42 & 60.61 & \underline{69.98} & 80.59 \\
LongLive~\cite{longlive}
& 97.39 & 96.32 & 98.78 & 41.17 & \underline{61.16} & 68.81 & 80.55 \\
Deep Forcing~\cite{deepforcing}
& 96.70 & 95.93 & 98.20 & \underline{46.93} & 59.80 & 68.81 & 80.53 \\
MemRoPE~\cite{memrope}
& \underline{97.74} & 96.32 & \underline{98.90} & 42.53 & 59.54 & 68.40 & 80.39 \\
\midrule
\textbf{FadeMem-TF}
& \textbf{97.86} & \textbf{96.55} & \textbf{99.03} & 42.19 & 60.98 & 69.62 & \underline{80.93} \\
\midrule
\multicolumn{8}{l}{\textit{With light FadeMem fine-tuning}} \\
\textbf{FadeMem-FT}
& 97.73 & \underline{96.50} & 98.85 & \textbf{47.06} & \textbf{61.41} & \textbf{70.70} & \textbf{81.59} \\
\bottomrule
\end{tabular}
\caption{
Quantitative comparison in the 60-second single-prompt setting.
We evaluate 128 MovieGenBench prompts with VBench-Long metrics; Avg. is the weighted average with half weight on dynamic degree.
FadeMem-TF denotes the inference-time memory variant without additional training, while FadeMem-FT denotes the lightly fine-tuned variant.
Best and second-best results are shown in bold and underlined.
}
\label{tab:long_video_results}
\end{table*}
\subsection{Experimental Setup}
\paragraph{Implementation Details.}
We implement FadeMem on Wan2.1-T2V-1.3B~\cite{wan2025wan} within the LongLive~\cite{longlive} autoregressive generation framework.
Videos are generated in three-latent-frame chunks with four denoising steps at timesteps $\{1000,750,500,250\}$.
Unless otherwise specified, we use $M=12$ historical entries and $\beta=0.3$, update memory after the final denoising step, and attend to 15 KV frame slots, including three from the current chunk.
\textbf{FadeMem-TF} modifies only inference-time memory organization, whereas 
\textbf{FadeMem-FT} is initialized from the same pretrained LongLive checkpoint and fine-tuned for only 1000 steps.
All methods use official implementations and matched evaluation settings.
\paragraph{Evaluation Protocol.}
Following prior work~\cite{longlive,memflow,selfforcing,cai2024pyramidkv}, we evaluate long-horizon text-to-video generation on prompts sampled from MovieGenBench~\cite{polyak2024moviegen}.
Experiments use 60-second videos at $480 \times 832$ resolution and 16 FPS; the long-duration evaluation extends the same protocol to 240 seconds.
We report VBench-Long~\cite{vbenchpp} metrics and Gemini 3.1-Pro visual stability scores following the Self-Forcing++ protocol~\cite{selfforcingpp}.
\subsection{Results}
\paragraph{Quantitative results.}
Table~\ref{tab:long_video_results} reports the quantitative comparison in the 60-second single-prompt setting.
Without additional training, FadeMem-TF achieves the highest subject consistency, background consistency, and motion smoothness, while raising the average score from 80.55 for LongLive~\cite{longlive} to 80.93.
FadeMem-TF prioritizes long-range consistency while maintaining a competitive Dynamic Degree of 42.19.
With light fine-tuning, FadeMem-FT raises Dynamic Degree to 47.06, slightly exceeding Deep Forcing's~\cite{deepforcing} 46.93, and reaches the highest average score of 81.59, improving over LongLive and MemFlow~\cite{memflow} by 1.04 and 1.00 points, respectively.
It combines the highest Dynamic Degree (47.06), Aesthetic Quality (61.41), and Imaging Quality (70.70).
These results show that FadeMem provides a stronger bounded-cache trade-off between long-range consistency, visual quality, and motion dynamics.


\paragraph{Long-duration results.}
To evaluate FadeMem over substantially longer horizons, we further evaluate 240-second generation under the same protocol.
As shown in Table~\ref{tab:long_duration}, FadeMem achieves the best aggregate consistency, Dynamic Degree, aggregate quality, and weighted average. This result shows that distance-aware memory consolidation remains effective as the amount of generated history continues to grow.

\begin{table}[t]
\centering
\small
\setlength{\tabcolsep}{1.2pt}
\renewcommand{\arraystretch}{1.05}
\begin{tabular}{lcccc}
\toprule
\textbf{Method} &
\makecell[c]{\textbf{Consistency}\\\textbf{Avg.} $\uparrow$} &
\makecell[c]{\textbf{Dynamic}\\\textbf{Degree} $\uparrow$} &
\makecell[c]{\textbf{Quality}\\\textbf{Avg.} $\uparrow$} &
\makecell[c]{\textbf{Weighted}\\\textbf{Avg.} $\uparrow$} \\
\midrule
MemFlow & 97.87 & 29.54 & \underline{67.42} & 80.58 \\
LongLive & 97.88 & 33.21 & 67.19 & \underline{80.84} \\
Deep Forcing & 96.96 & \underline{39.17} & 65.28 & 80.18 \\
MemRoPE & \underline{97.94} & 38.25 & 65.36 & 80.67 \\
\midrule
\textbf{FadeMem (Ours)} & \textbf{98.03} & \textbf{41.13} & \textbf{68.64} & \textbf{82.17} \\
\bottomrule
\end{tabular}
\caption{Aggregated VBench-Long results for 240-second generation.
Consistency Avg. is the arithmetic mean of Subject Consistency, Background Consistency, and Motion Smoothness; Quality Avg. is the arithmetic mean of Aesthetic Quality and Imaging Quality.
Weighted Avg. follows Table~\ref{tab:long_video_results}.
Best and second-best results are shown in bold and underlined.}
\label{tab:long_duration}
\end{table}


\paragraph{Efficiency.}
Table~\ref{tab:efficiency} compares efficiency under a matched 15-slot KV budget using BF16 with batch size 1 on a single RTX PRO 6000 GPU. 
Core FPS excludes text encoding, VAE decoding, and disk I/O. FadeMem achieves the lowest peak memory usage (22.72 GiB) while
maintaining throughput close to MemRoPE (11.24 versus 11.85 FPS).

\begin{table}[t]
\centering
\small
\setlength{\tabcolsep}{3.5pt}
\renewcommand{\arraystretch}{1.00}
\begin{tabular}{lccc}
\toprule
\textbf{Method} &
\makecell[c]{\textbf{Attention}\\\textbf{KV Slots}} &
\makecell[c]{\textbf{Core FPS}\\$\uparrow$} &
\makecell[c]{\textbf{Peak Alloc.}\\\textbf{Memory (GiB)} $\downarrow$} \\
\midrule
LongLive & 15 & \textbf{15.78} & 30.51 \\
MemRoPE & 15 & 11.85 & 25.84 \\
\textbf{FadeMem (Ours)} & 15 & 11.24 & \textbf{22.72} \\
\bottomrule
\end{tabular}
\caption{Efficiency comparison under a matched attention KV budget.}
\label{tab:efficiency}
\end{table}


\paragraph{VLM-based evaluation.}
To complement VBench-Long metrics, we further report VLM-based visual stability scores in the 60-second setting using Gemini 3.1-Pro, following the evaluation protocol of Self-Forcing++~\cite{selfforcingpp}.
As shown in Table~\ref{tab:vlm_evaluation}, FadeMem attains the highest score of 4.84, providing complementary evidence of its long-range visual stability.
\begin{table}[t]
\centering
\small
\setlength{\tabcolsep}{10pt}
\renewcommand{\arraystretch}{1.05}
\begin{tabular}{l c}
\toprule
\textbf{Method} & \textbf{Stability} $\uparrow$ \\
\midrule
MemFlow~\cite{memflow} & 4.77 \\
LongLive~\cite{longlive} & 4.74 \\
Deep Forcing~\cite{deepforcing} & 4.51 \\
MemRoPE~\cite{memrope} & 4.80 \\
\midrule
\textbf{FadeMem (Ours)} & \textbf{4.84} \\
\bottomrule
\end{tabular}
\caption{Visual stability (VLM).
We report Gemini 3.1-Pro visual stability scores in the 60-second setting following the Self-Forcing++ evaluation protocol.
Higher scores indicate stronger long-range visual stability.}
\label{tab:vlm_evaluation}
\end{table}
\begin{figure*}[t]
    \centering
    \includegraphics[width=\textwidth]{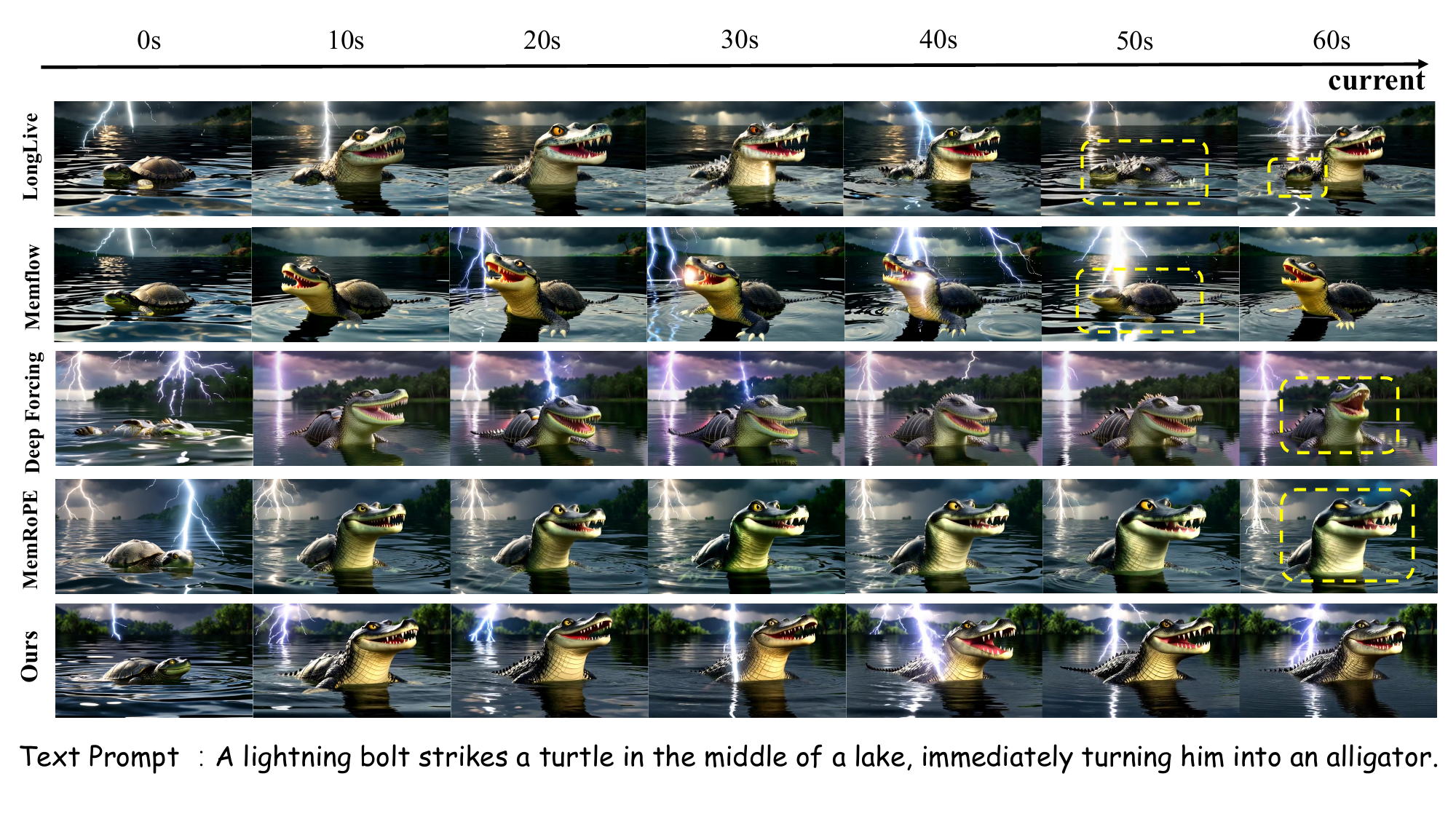}
    \caption{Qualitative comparison in the 60-second setting for a prompt requiring an early turtle-to-alligator transition. Yellow boxes highlight representative late-stage regions.}
    \label{fig:long_video_qual}
\end{figure*}


\paragraph{Qualitative results.}
Figure~\ref{fig:long_video_qual} evaluates whether each method retains an early prompt-induced semantic transition over a long rollout. LongLive~\cite{longlive} and MemFlow~\cite{memflow} later recover turtle-like cues, whereas Deep Forcing~\cite{deepforcing} and MemRoPE~\cite{memrope} preserve the coarse alligator identity. FadeMem further maintains sharper subject details, richer water-lighting interactions, and a more coherent stormy appearance, demonstrating stronger long-range identity preservation and local visual quality under a bounded KV budget.

\subsection{Ablation Studies}
We ablate three design choices in \textbf{FadeMem}: the distance-warping exponent, the local consolidation operator, and the first-frame global anchor.
All ablations use the 60-second training-free setting to isolate each memory design choice.

\begin{table*}[t!]
\centering
\small
\setlength{\tabcolsep}{0.8pt}
\renewcommand{\arraystretch}{1.00}
\begin{tabular}{llccccccc}
\toprule
\multirow{2}{*}{\textbf{Study}} &
\multirow{2}{*}{\textbf{Variant}} &
\multicolumn{7}{c}{\textbf{Evaluation Scores} $\uparrow$} \\
\cmidrule(lr){3-9}
&
&
\makecell[c]{Subject\\Consistency} &
\makecell[c]{Background\\Consistency} &
\makecell[c]{Motion\\Smoothness} &
\makecell[c]{Dynamic\\Degree} &
\makecell[c]{Aesthetic\\Quality} &
\makecell[c]{Imaging\\Quality} &
\makecell[c]{Avg.} \\
\midrule
\multirow{5}{*}{\makecell[l]{Temporal\\allocation}}
& $\beta=0.1$ & \underline{97.82} & \underline{96.49} & \underline{99.02} & 42.84 & 60.55 & 69.05 & 80.79 \\
& $\beta=0.3$ & \textbf{97.86} & \textbf{96.55} & \textbf{99.03} & 42.19 & \textbf{60.98} & 69.62 & 80.93 \\
& $\beta=0.5$ & 97.78 & 96.48 & 98.99 & 44.04 & 60.72 & 69.71 & 81.04 \\
& $\beta=0.7$ & 97.81 & 96.44 & 98.96 & \textbf{44.48} & \underline{60.88} & \textbf{70.18} & \textbf{81.18} \\
& $\beta=0.9$ & 97.76 & 96.41 & 98.95 & \underline{44.09} & 60.62 & \underline{70.01} & \underline{81.05} \\
\midrule
\multirow{4}{*}{\makecell[l]{Consolidation\\operator}}
& Select Nearest & 97.76 & 96.41 & 98.96 & \textbf{44.79} & 60.21 & 68.19 & 80.71 \\
\addlinespace[0.15em]
& Average & 97.82 & 96.48 & \underline{99.01} & \underline{43.39} & \underline{60.64} & \underline{69.36} & \underline{80.91} \\
\addlinespace[0.15em]
& Weighted Average & \textbf{97.86} & \underline{96.55} & \textbf{99.03} & 42.19 & \textbf{60.98} & \textbf{69.62} & \textbf{80.93} \\
\addlinespace[0.15em]
& Max Pooling & \underline{97.83} & \textbf{96.71} & 98.60 & 11.80 & 46.67 & 57.72 & 73.35 \\
\midrule
\multirow{2}{*}{\makecell[l]{Global\\anchor}}
& \makecell[l]{w/o First-frame\\Anchor} & \textbf{97.99} & \textbf{96.72} & 99.01 & 33.80 & 60.88 & \textbf{70.24} & 80.32 \\
& \makecell[l]{w/ First-frame\\Anchor} & 97.86 & 96.55 & \textbf{99.03} & \textbf{42.19} & \textbf{60.98} & 69.62 & \textbf{80.93} \\
\bottomrule
\end{tabular}
\caption{
Ablation studies in the 60-second setting.
Rows are grouped by design factor; best results are bold, and second-best results are underlined for groups with more than two variants.
Avg. is the weighted average with half weight on dynamic degree.
}
\label{tab:ablation_beta}
\label{tab:ablation_consolidation}
\label{tab:ablation_global_anchor}
\end{table*}



\paragraph{Temporal allocation exponent.}
The exponent $\beta$ balances dense recent context and broader distant coverage. As shown in Table~\ref{tab:ablation_beta}, $\beta=0.3$ achieves the best subject consistency, background consistency, motion smoothness, and aesthetic quality, whereas $\beta=0.7$ favors Dynamic Degree, Imaging Quality, and the weighted average. We use $\beta=0.3$ to prioritize stable long-horizon identity and scene preservation while retaining competitive overall performance.

\paragraph{Memory consolidation operator.}
We next ablate how FadeMem merges the adjacent pair selected by the temporal allocation schedule. As shown in
Table~\ref{tab:ablation_consolidation}, averaging-based consolidation provides a better overall balance than representative selection or max pooling. Select Nearest favors motion dynamics but weakens consistency and visual quality, while Max Pooling substantially degrades both, suggesting that hard aggregation over-compresses the cached states. Average and Weighted Average perform similarly, with Weighted Average achieving the best overall score; we therefore use it by default.

\paragraph{First-frame global anchor.}
Finally, we ablate whether FadeMem should preserve the first frame as a global anchor.
As shown in Table~\ref{tab:ablation_global_anchor}, removing the anchor slightly improves some consistency metrics but sharply reduces dynamic degree.
This suggests overly conservative rollouts with limited temporal evolution.
Keeping the anchor provides a better trade-off between global coherence and motion progression, so we use it by default.


\section{Conclusion}
We present \textbf{FadeMem}, a distance-aware memory consolidation mechanism for long-horizon autoregressive video generation under a fixed KV cache budget. Motivated by distance-dependent spectral decay, FadeMem replaces manually partitioned cache designs with a unified ordered memory whose temporal resolution fades with distance. Recent entries are preserved at fine granularity for local dynamics, while distant entries are progressively merged into coarser span-level anchors for scene layout, appearance, and identity. Experiments show that this simple cache organization improves long-range consistency without architectural changes. These results suggest that long-video coherence depends not only on how much history is retained, but also on whether history is represented at an appropriate temporal granularity as it recedes into the past.


\section{Limitations}
\textbf{FadeMem} uses a fixed allocation schedule that may be suboptimal for abrupt scene transitions, fast motion, frequent semantic changes, or detail-sensitive objects. Stochastic initial noise can occasionally produce transient motion stalls, although the rollout may recover as these states are progressively consolidated and the prompt and first-frame anchor regain influence. FadeMem also inherits limitations of the base generator, including prompt misalignment, implausible physical dynamics, weak action planning, and semantic drift.

\clearpage
\section*{Supplementary Material}
\addcontentsline{toc}{section}{Supplementary Material}
\setcounter{section}{0}
\setcounter{figure}{0}
\setcounter{table}{0}
\setcounter{algorithm}{0}
\renewcommand{\thesection}{S\arabic{section}}
\renewcommand{\thefigure}{S\arabic{figure}}
\renewcommand{\thetable}{S\arabic{table}}
\renewcommand{\thealgorithm}{S\arabic{algorithm}}

\section{FadeMem Memory Update}
\label{app:algorithm}


Algorithm~\ref{alg:fademem_update} gives the complete online update used by FadeMem.
The eligible-pair set excludes the pair containing the just-inserted newest entry.
When the global-anchor option is enabled, the first-frame anchor remains eligible for pair selection, but its original KV block and temporal position are retained whenever its pair is selected.

\begin{algorithm}[H]
\caption{FadeMem Memory Update}
\label{alg:fademem_update}
\footnotesize
\begin{algorithmic}[1]
\Require Memory $\mathcal{M}_{t-1}$; new KV block
$\mathbf{B}_t=(\mathbf{K}_t,\mathbf{V}_t)$; budget $M$; exponent $\beta$
\Ensure Updated memory $\mathcal{M}_t$

\State $\mathcal{M}_t \gets
\operatorname{Append}
\left(
\mathcal{M}_{t-1},
(\mathbf{K}_t,\mathbf{V}_t,t,1)
\right)$

\If{$|\mathcal{M}_t| \leq M$}
    \State \Return $\mathcal{M}_t$
\EndIf


\For{each adjacent pair $(m_i,m_{i+1})$ except the pair containing the newest entry}
    \State $g_i \gets \left|(t-\mu_{i+1})^\beta - (t-\mu_i)^\beta\right|$
\EndFor

\State $j \gets \arg\min_i g_i$
\State $s_{\mathrm{new}} \gets s_j+s_{j+1}$

\If{the selected pair contains the first-frame anchor $m_a$}
    \State $\mu_{\mathrm{new}} \gets 0$
    \State $\bar{\mathbf{K}}_{\mathrm{new}}
    \gets \bar{\mathbf{K}}_a$
    \State $\bar{\mathbf{V}}_{\mathrm{new}}
    \gets \bar{\mathbf{V}}_a$
\Else
    \State $\mu_{\mathrm{new}} \gets
    (s_j\mu_j+s_{j+1}\mu_{j+1})/s_{\mathrm{new}}$

    \State $w_j \gets
    \sqrt{s_j}/(\sqrt{s_j}+\sqrt{s_{j+1}})$
    \State $w_{j+1} \gets 1-w_j$

    \State $\bar{\mathbf{K}}_{\mathrm{new}}
    \gets
    w_j\bar{\mathbf{K}}_j+
    w_{j+1}\bar{\mathbf{K}}_{j+1}$

    \State $\bar{\mathbf{V}}_{\mathrm{new}}
    \gets
    w_j\bar{\mathbf{V}}_j+
    w_{j+1}\bar{\mathbf{V}}_{j+1}$
\EndIf

\State Replace $(m_j,m_{j+1})$ with
$(\bar{\mathbf{K}}_{\mathrm{new}},
\bar{\mathbf{V}}_{\mathrm{new}},
\mu_{\mathrm{new}},s_{\mathrm{new}})$

\State \Return $\mathcal{M}_t$
\end{algorithmic}
\end{algorithm}



\begin{figure*}[t]
    \centering
    \includegraphics[width=\textwidth]{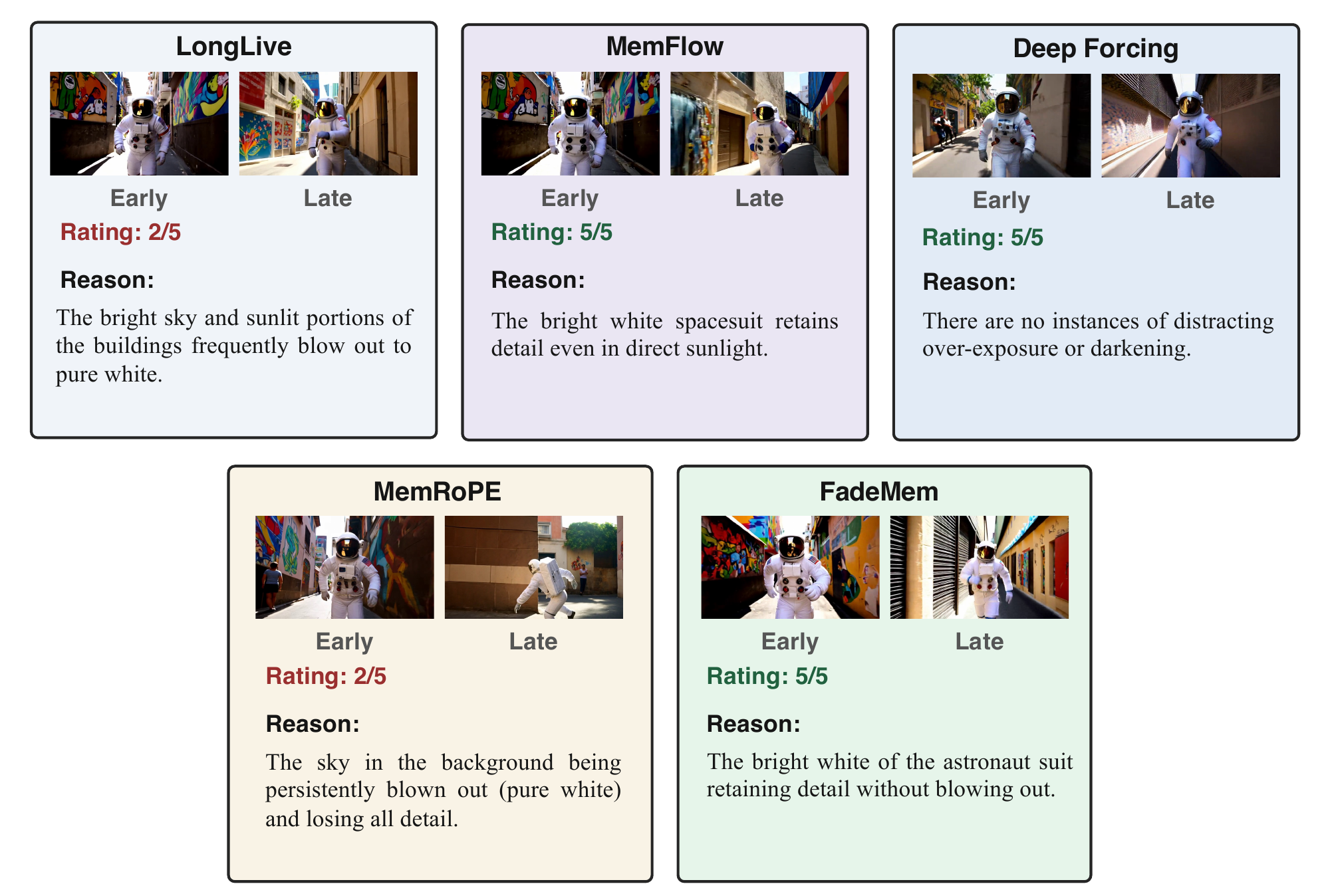}
    \caption{Example VLM evaluation for exposure stability.
    Each panel shows frames near the beginning (1 second) and end (58 seconds) of a 60-second video generated from the same prompt.
    Ratings and abridged reasoning excerpts are taken from the corresponding Gemini 3.1-Pro Preview response to each complete video.}
    \label{fig:vlm_evaluation_example}
\end{figure*}

\section{Spectral Analysis Details}
\label{app:spectral_analysis}

\paragraph{Data and preprocessing.}
We collect 254 valid natural-video files from Pexels and resize them to $832 \times 480$ resolution.
Long videos are divided into non-overlapping clips of 708 RGB frames, yielding 374 fixed-length clips in total.
The native frame rates are retained during preprocessing; therefore, temporal distances in this analysis are reported in latent-frame steps rather than seconds.

\paragraph{Latent frequency decomposition.}
We encode each clip using the Wan2.1 VAE distributed with the Wan2.1-T2V-1.3B checkpoint~\cite{wan2025wan}.
Each clip produces 177 latent frames with 16 channels at a spatial resolution of $60 \times 104$.
We average the latent channels and apply an orthonormal two-dimensional discrete cosine transform (DCT) to each latent frame.
For a DCT coefficient at spatial-frequency coordinate $(u,v)$, we define its normalized radial frequency as
\[
r(u,v)=\sqrt{(u/H)^2+(v/W)^2},
\]
where $H$ and $W$ denote the spatial dimensions of the latent feature.

The DCT coefficients are grouped according to their normalized radial frequencies.
For visualization, the spectral-analysis figure in the main paper aggregates them into low-, middle-, and high-frequency ranges in panel (a), while panel (b) uses a finer radial partition to estimate the stable frequency boundary.

\paragraph{Temporal correlation and stable bandwidth.}
Let $D(z_t)$ denote the DCT coefficients of latent frame $z_t$, and let $\mathcal{B}_k$ denote a radial frequency band.
For a temporal lag $\Delta$, we compute
\[
C_k(\Delta)=
\operatorname{Pearson}
\left(
D(z_0)|_{\mathcal{B}_k},
D(z_{\Delta})|_{\mathcal{B}_k}
\right),
\]
where the DCT coefficients within each frequency band are treated as a vector.
We average the resulting correlations over all 374 clips.
Panel (a) of the spectral-analysis figure in the main paper reports the mean correlation across the frequency plane, together with the aggregated low-, middle-, and high-frequency correlations.

For each radial frequency band, we smooth its cross-video correlation curve and identify its decorrelation horizon $t_k^*$ as the onset of the near-flat regime following the main correlation decay.
Associating each $t_k^*$ with the center radius of its frequency band produces the boundary points in panel (b).
We then fit the relationship between temporal distance and stable frequency radius using
\[
r^*(t)=a t^{-b}
\]
with nonlinear least squares.
The resulting trend indicates that the range of stable frequencies decreases approximately according to a power law as temporal distance increases.

This empirical observation motivates the power-law form of FadeMem's temporal allocation.
The memory-allocation exponent $\beta$ is selected empirically through the ablation in the main paper, rather than being set directly to the fitted spectral exponent, because stable frequency bandwidth and memory-entry density represent different quantities.

\section{VLM Evaluation Details}
\label{app:vlm_evaluation}

\paragraph{Evaluation protocol.}
The VLM-based score in the main paper specifically measures exposure stability.
Following recent long-video work that uses VLM judges to assess exposure quality~\cite{selfforcingpp,memrope}, we use Gemini 3.1-Pro Preview (\texttt{gemini-3.1-pro-preview}) as the evaluator.
For each method, each of the 128 complete 60-second videos is submitted to the evaluator once with the temperature set to 0.
The evaluator receives the complete generated video and the exposure-stability rubric below, but not the corresponding generation prompt.
All 128 requests complete successfully for every evaluated method.
We retain the raw model responses, parse the scalar rating returned for each video, and report the arithmetic mean over the 128 videos.

\paragraph{Scoring rubric.}
The following instruction is provided to the evaluator:

\begin{quote}
You are tasked with rating the exposure stability of a video.
Assign a score according to the following scale:

\textbf{0: Catastrophic Exposure.}
Nearly the entire frame is either blown out (pure white) or crushed (pure black), rendering the scene unreadable.

\textbf{1: Severe Exposure Failure.}
Large portions are dominated by over- or under-exposure, substantially impairing visibility.

\textbf{2: Noticeable Exposure Problems.}
Persistent clipping in highlights or shadows; significant areas lose detail.

\textbf{3: Moderate Exposure Issues.}
Over-exposed highlights or under-exposed shadows occur but are limited in extent or duration.

\textbf{4: Minor Exposure Flaws.}
Small regions are occasionally too bright or too dark, but do not meaningfully disrupt visibility.

\textbf{5: Well-Exposed.}
Balanced lighting across the frame with no distracting over-exposure or darkening.
\end{quote}

\paragraph{Qualitative example.}
Figure~\ref{fig:vlm_evaluation_example} illustrates how the rubric is applied to one shared generation prompt.
The early and late frames are shown only as a concise visualization for readers; the evaluator receives and assesses each complete 60-second video rather than these two frames alone.

\section{Implementation and Reproducibility Notes}
\label{app:implementation}

We implement FadeMem on Wan2.1-T2V-1.3B~\cite{wan2025wan} within the LongLive autoregressive generation framework~\cite{longlive}.
Videos are generated in three-latent-frame chunks with four denoising steps at timesteps $\{1000,750,500,250\}$.
Each memory entry stores the KV block of one generated latent frame.
The default configuration uses $M=12$ historical entries and three current latent-frame slots, for 15 KV frame slots per attention computation, with $\beta=0.3$; memory is updated after the final denoising step.
The training-free variant changes only inference-time memory organization.
The fine-tuned variant follows LongLive's second-stage streaming-training setup, updates only LoRA adapters, and keeps the pretrained backbone fixed.

The primary evaluation uses 128 MovieGenBench prompts, 60-second videos at $480 \times 832$ resolution and 16 FPS, and VBench-Long metrics.
The long-duration evaluation extends the same setup to 240 seconds.
Efficiency is measured with batch size 1 and BF16 precision on a single NVIDIA RTX PRO 6000 Blackwell Server Edition GPU.
Core FPS excludes text encoding, VAE decoding, and disk I/O; peak allocated memory is the maximum allocated by PyTorch during a generation request.

\section{Additional Qualitative Results}
\label{app:additional_qualitative}

We provide additional qualitative results covering diverse 60-second
generations, a comparison between the training-free and lightly tuned
variants, and continuous rollouts extending to 240 seconds and one hour.
All frames are sampled in temporal order from their corresponding
generated videos.

\begin{figure*}[t]
    \centering
    \includegraphics[width=\textwidth]
    {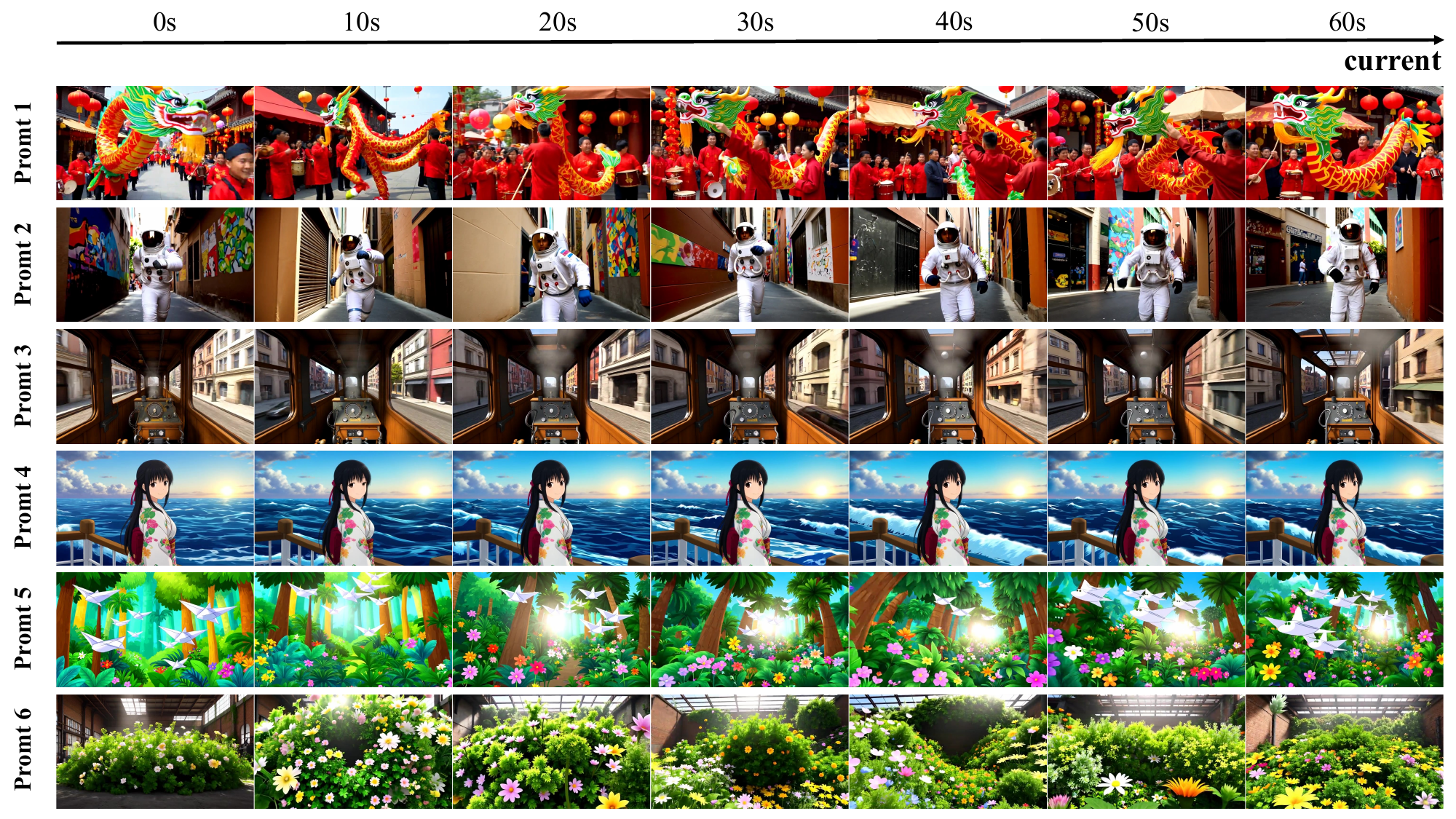}
    \caption{
    Additional 60-second qualitative results of FadeMem.
    Each row corresponds to one continuous generated video and shows
    frames sampled at 10-second intervals. Across diverse subjects,
    scenes, and motion patterns, FadeMem preserves recognizable visual
    content and coherent scene evolution throughout the generation.
    }
    \label{fig:additional_60s}
\end{figure*}

\paragraph{Additional 60-second examples.}
Figure~\ref{fig:additional_60s} presents six additional examples covering
diverse subjects, scenes, and motion patterns. These results further
demonstrate the ability of FadeMem to preserve recognizable visual
content and coherent scene evolution throughout long generation
sequences.

\begin{figure*}[t]
    \centering
    \includegraphics[width=\textwidth]
    {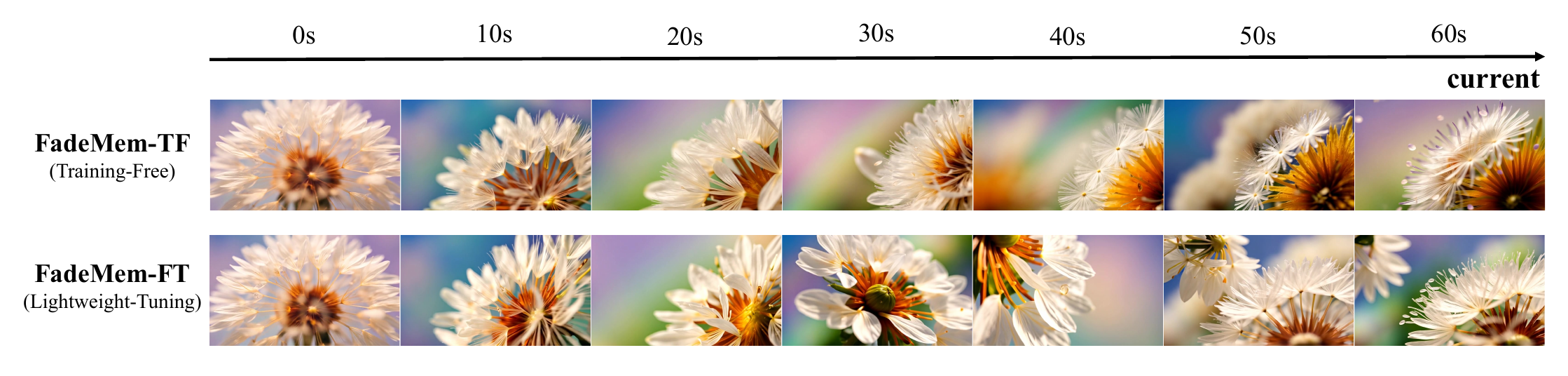}
    \caption{
    Qualitative comparison between FadeMem-TF (Training-Free) and
    FadeMem-FT (Light-Tuning) at matched timestamps. Training-Free
    preserves the dandelion seed-head semantics through the macro zoom,
    while Light-Tuning provides a smoother visual trajectory and higher
    visual fidelity.
    }
    \label{fig:tf_vs_ft}
\end{figure*}

\paragraph{Training-free versus light-tuning.}
Figure~\ref{fig:tf_vs_ft} compares the two FadeMem variants at matched
timestamps. The comparison illustrates their complementary behaviors
during long-horizon generation.

\begin{figure*}[t]
    \centering
    \includegraphics[width=\textwidth]
    {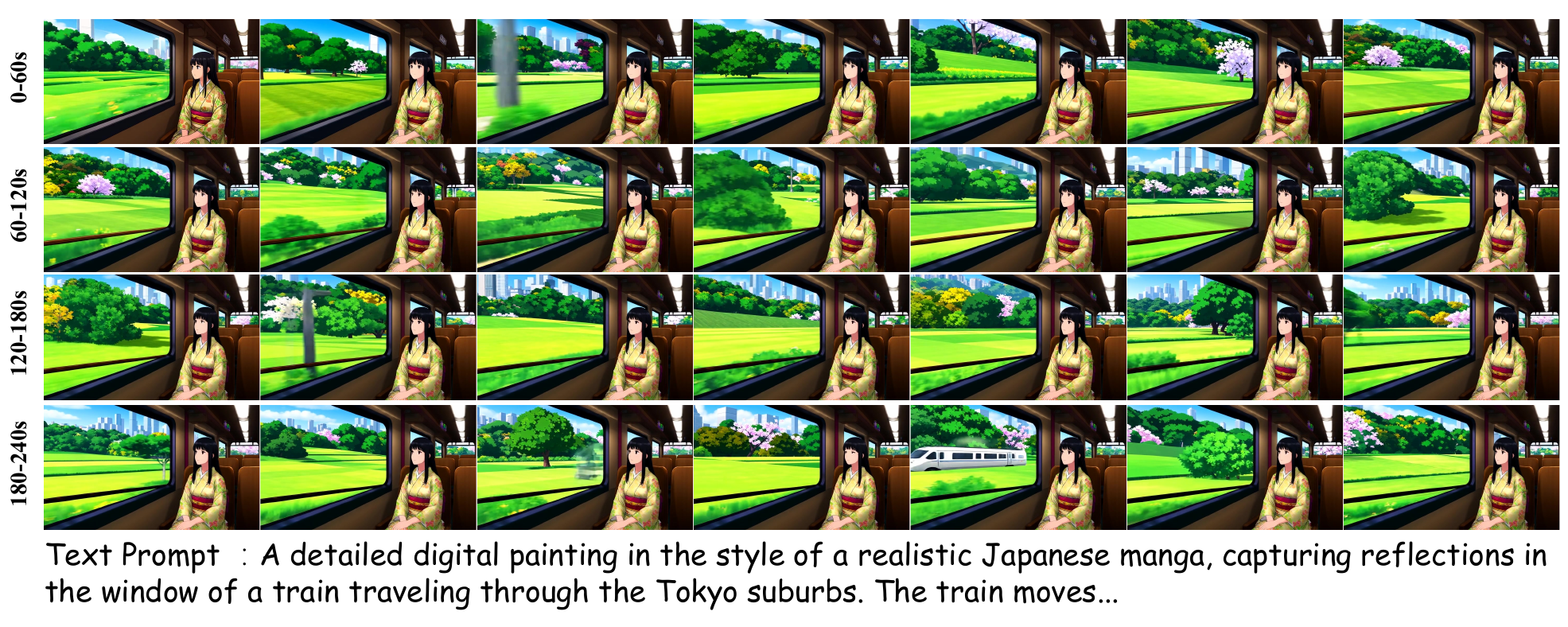}
    \caption{
    Representative frames from one continuous 240-second FadeMem
    generation. Each row covers one consecutive 60-second interval,
    with frames arranged in temporal order. Across the full four-minute
    rollout, FadeMem consistently preserves the subject identity,
    attire, train interior, and surrounding scene while maintaining
    natural and meaningful temporal evolution.
    }
    \label{fig:additional_240s}
\end{figure*}

\paragraph{Continuous 240-second generation.}
Figure~\ref{fig:additional_240s} provides a dense visualization of one
continuous four-minute rollout. The four rows cover consecutive
60-second intervals, showing how the generated content evolves over a
substantially longer temporal horizon.

\begin{figure*}[t]
    \centering
    \includegraphics[width=\textwidth]
    {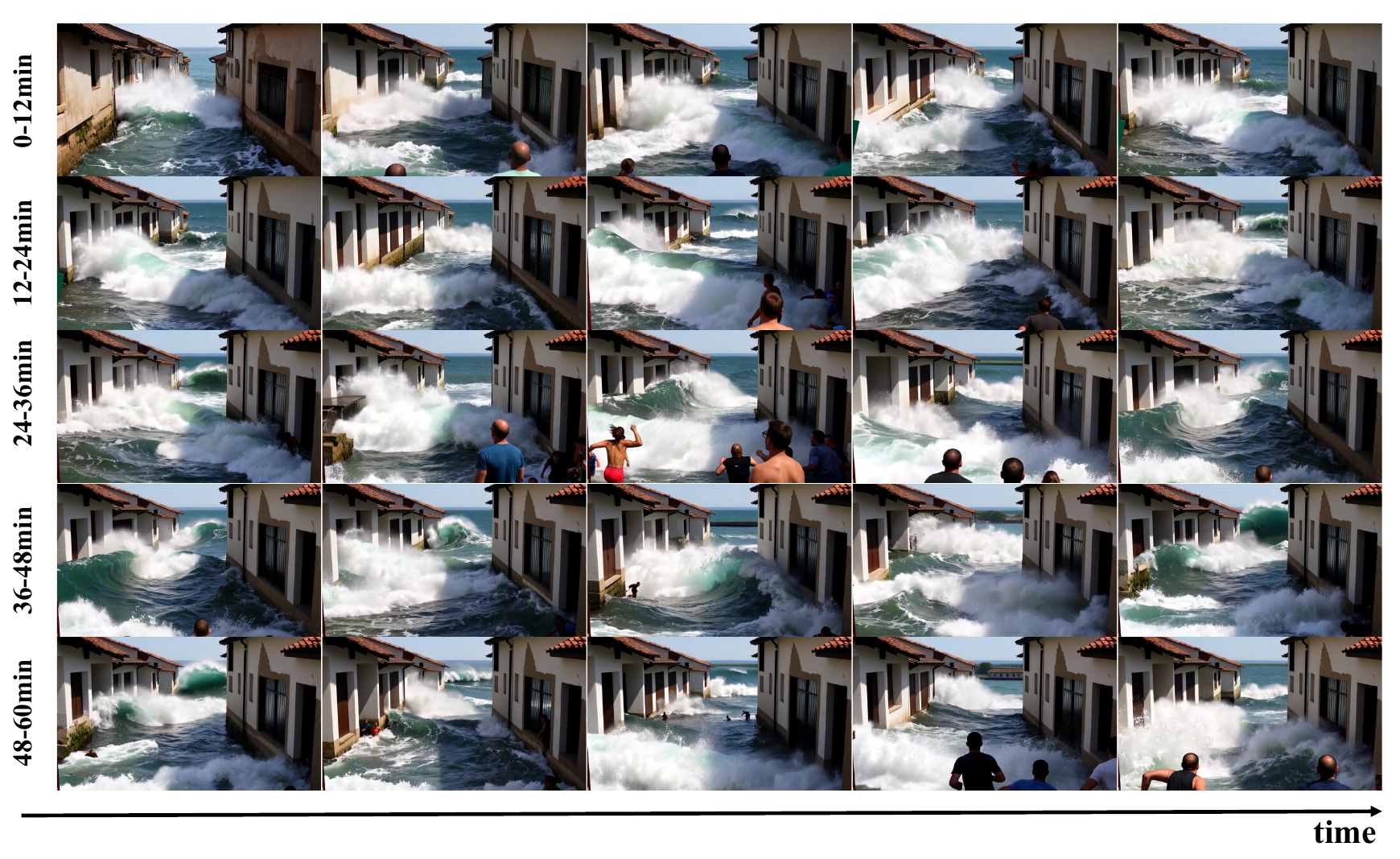}
    \caption{
    Representative frames from one continuous one-hour FadeMem
    generation under a fixed cache budget. Each row covers one
    consecutive 12-minute interval. FadeMem maintains the recognizable
    narrow-alley structure and intense tsunami dynamics throughout the
    hour-long rollout.
    }
    \label{fig:additional_1hour}
\end{figure*}

\paragraph{Continuous one-hour generation.}
Figure~\ref{fig:additional_1hour} presents an hour-scale generation
under a fixed cache budget, providing a qualitative stress test over an
extremely long generated history.

\bibliography{references}
\end{document}